\documentclass[10pt,twocolumn,letterpaper]{article}

\usepackage{times}
\usepackage{epsfig}
\usepackage{graphicx}
\usepackage{amsmath}
\usepackage{amssymb}
\usepackage{float}
\restylefloat{table}
\usepackage[margin=0.75in]{geometry}

\usepackage[utf8]{inputenc}
\usepackage{enumitem}
\usepackage{subfigure}
\usepackage{graphics}

\usepackage{supertabular}

\usepackage[pagebackref=true,breaklinks=true,letterpaper=true,colorlinks,bookmarks=false]{hyperref}

\usepackage{fancyhdr}
\usepackage{lipsum} 

\fancypagestyle{plain}{
  \fancyhf{}
  \fancyfoot[C]{\thepage}%
}
\fancypagestyle{firstpage}{
  \fancyhf{}
  \fancyfoot[L]{$^{*}$ This work was done when the authors were at University of California, San Diego. $^{+}$The authors can be reached at snagesh@ucsd.edu, asbaig@ucsd.edu, sasriniv@ucsd.edu, arangesh@ucsd.edu, mtrivedi@ucsd.edu}%
}

\begin{document}

    \title{\textbf{Structure Aware and Class Balanced 3D Object Detection on nuScenes Dataset}}

\author{Sushruth Nagesh$^{*+}$, Asfiya Baig$^{*+}$, Savitha Srinivasan$^{*+}$, Akshay Rangesh$^{+}$, Mohan Trivedi$^{+}$\\
UC San Diego\\

}
\date{}
\maketitle
\thispagestyle{firstpage}
\begin{abstract}
   3-D object detection is pivotal for autonomous driving. Point cloud based methods have become increasingly popular for 3-D object detection, owing to their accurate depth information. NuTonomy's nuScenes dataset \cite{nuscenes} greatly extends commonly used datasets such as KITTI in size, sensor modalities, categories, and annotation numbers. However, it suffers from severe class imbalance. The Class-balanced Grouping and Sampling paper \cite{det3d} addresses this issue and suggests augmentation and sampling strategy. However, the localization precision of this model is affected by the loss of spatial information in the downscaled feature maps. We propose to enhance the performance of the CBGS model by designing an auxiliary network \cite{sassd}, that makes full use of the structure information of the 3D point cloud, in order to improve the localization accuracy. The detachable auxiliary network is jointly optimized by two point-level supervisions, namely foreground segmentation and center estimation. The auxiliary network does not introduce any extra computation during inference, since it can be detached at test time. 
\end{abstract}

\section{Introduction}
3D object detection serves as an essential basis of visual perception, motion prediction, and planning for autonomous driving. Many of the current methods rely on point cloud data for accurate 3D detection. Datasets such as KITTI, which are widely used for this application have several drawbacks. They are limited in size, and have fewer categories that are not entirely representative of real-world scenarios. Many new datasets and testbeds like LISA-A \cite{lisatestbed}, LISA-Audi\cite{lisaaudi}, LISA-T \cite{oct}, Waymo\cite{waymo}, nuScenes\cite{nuscenes} etc have contributed significantly in terms of data volumes and complexities. 

Out of these datasets, nuScenes dataset has the largest number of lidar sweeps and it is collected under different environment conditions. nuScenes dataset has 10 categories for 3D detection as compared to 3 in KITTI, and also has 360 degree coverage in both vision and LIDAR modalities. 360 degree full-surrond data is important for reliable object detection and tracking as shown by Rangesh et al \cite{Nobs} where a full-surround camera and LiDAR based approach for tracking multiple objects in the autonomous driving context is developed. The 3D ground truth boxes also have attributes such as velocities associated with them, that aids in the classification process. 
However, a major drawback of the nuScenes dataset is its inherent class imbalance problem. The number of examples for common object classes  outnumbers that of rare object classes by a large margin. To address this issue, the CBGS \cite{det3d} paper proposes grouping and sampling strategies. It generates a smoother class distribution by improving the average density of rare classes in the training split.
\begin{figure}[h!]
  \includegraphics[width=9cm,height=4.2cm]{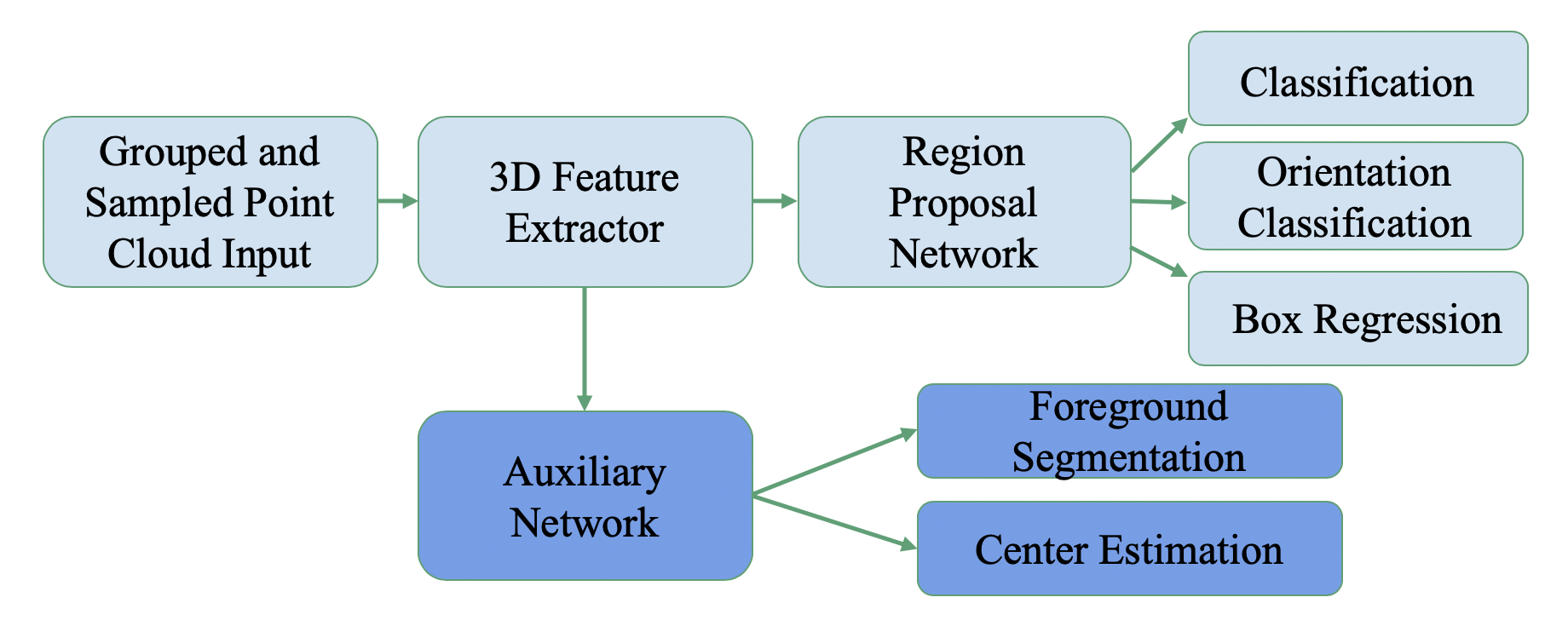}
  \caption{Struture Aware and Class Balanced 3D object detection overview}
\end{figure}

Although CBGS addresses the class imbalance problem, the network architecture does not utilize structure information of point cloud data, which leaves room for improvement in the model's localization performance. To enhance the model's localization precision, we propose to include an auxiliary network \cite{sassd}. The network exploits fine-grained structure information of point cloud by converting convolutional features back to point level representations. This is performed in order to learn object-sensitive boundaries and intra-object relationships. This additional network does not add to the computational overhead during inference, since it can it can be removed while testing. The auxiliary network performs two main tasks- foreground segmentation and center estimation. The segmentation task enables the backbone network to more precisely detect the object boundary. The center estimation task aids in learning the relative position of each object point with respect to the object center, leading to more accurate localization.

The main contributions of this paper can be summarized as below: 
\begin{itemize}
\item Introducing an auxiliary network into the existing CBGS architecture to improve localization accuracy. 

\item Introducing multi-group Non-Maximal Suppression (NMS) in the multi-group head network to decrease the number of false positive bounding boxes.
\end{itemize}







\section{Related Work}
\subsection{3D Object Detection} 
There are several methods proposed to tackle the task of 3D object detection. Rangesh et al \cite{ground} perform 3D object detection using single monocular images by leveraging the ground plane. Some works like \cite{3d_obj}, \cite{PIXOR}, \cite{pointpillars} convert point clouds to the bird-eye-view format, following which a 2D CNN is applied to obtain the 3D detection results. 
Some methods convert the point cloud into voxels \cite{amodal}, \cite{Second}, \cite{voxelnet} and apply 3D feature extractors using 3D Sparse Convolutions. Several Point-based methods first predict 2D boxes from the image, and then estimate the location, size and orientation of the 3D objects using the cropped point cloud by applying PointNet++ \cite{pointnet++}. However, most of the above methods don't perform well on all categories, especially the rare classes such as bicycle. Therefore, the class imbalance problem needs to be addressed in a more effective manner. Zhu et al \cite{det3d} address the class imbalance by designing a class-balanced sampling and augmentation strategy to generate a well balanced data distribution. Moreover, they use a balanced grouping head to improve the performance for categories with similar shapes and dataset sizes.
\subsection{Auxiliary Task Learning}
In the domain of autonomous driving, researchers have proposed to incorporate additional modules providing contextual information to boost the detection accuracy. The authors of SA-SSD \cite{sassd} propose to improve the localization accuracy  of single-stage detectors by exploiting the structure information of 3D point cloud using an auxiliary network.  Yang et al \cite{hdmaps} suggested to extract geometric and semantic features from High Definition (HD) Maps to enhance the network's awareness about road geometry . Liang et al \cite{featfusion} proposed feature fusion method  which combines information coming from multiple scales and 3D space. Zhao et al \cite{crowd}
proposed to improve the crowd counting by leveraging heterogeneous attributes of the density map as guidance to fully utilize the potential of the underlying representations without explicitly changing the extracted features.  
Mordan et al \cite{mordan} suggested to use auxiliary supervision to learn the scene-aware feature for improving the robustness of detecting objects. Our approach is mainly motivated by the auxiliary network suggested by \cite{sassd}, however differ from them at the same time as we are augmenting this into a two stage network\cite{det3d} and using nuscenes dataset\cite{nuscenes} 

\begin{figure}[h!]
\includegraphics[width=9cm,height=4.5cm]{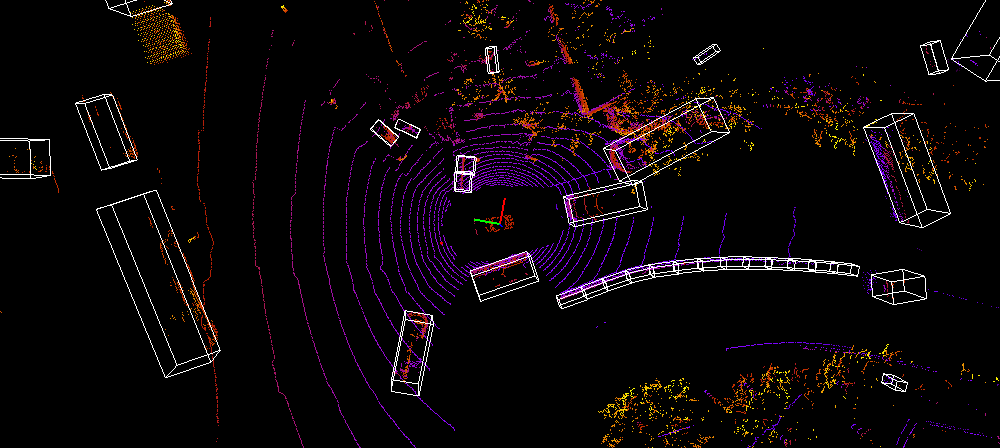}
  \caption{Example of a point cloud with ground truth boxes}
\end{figure}

\begin{figure*}[h!]
  \includegraphics[width=17.5cm,height=8.5cm]{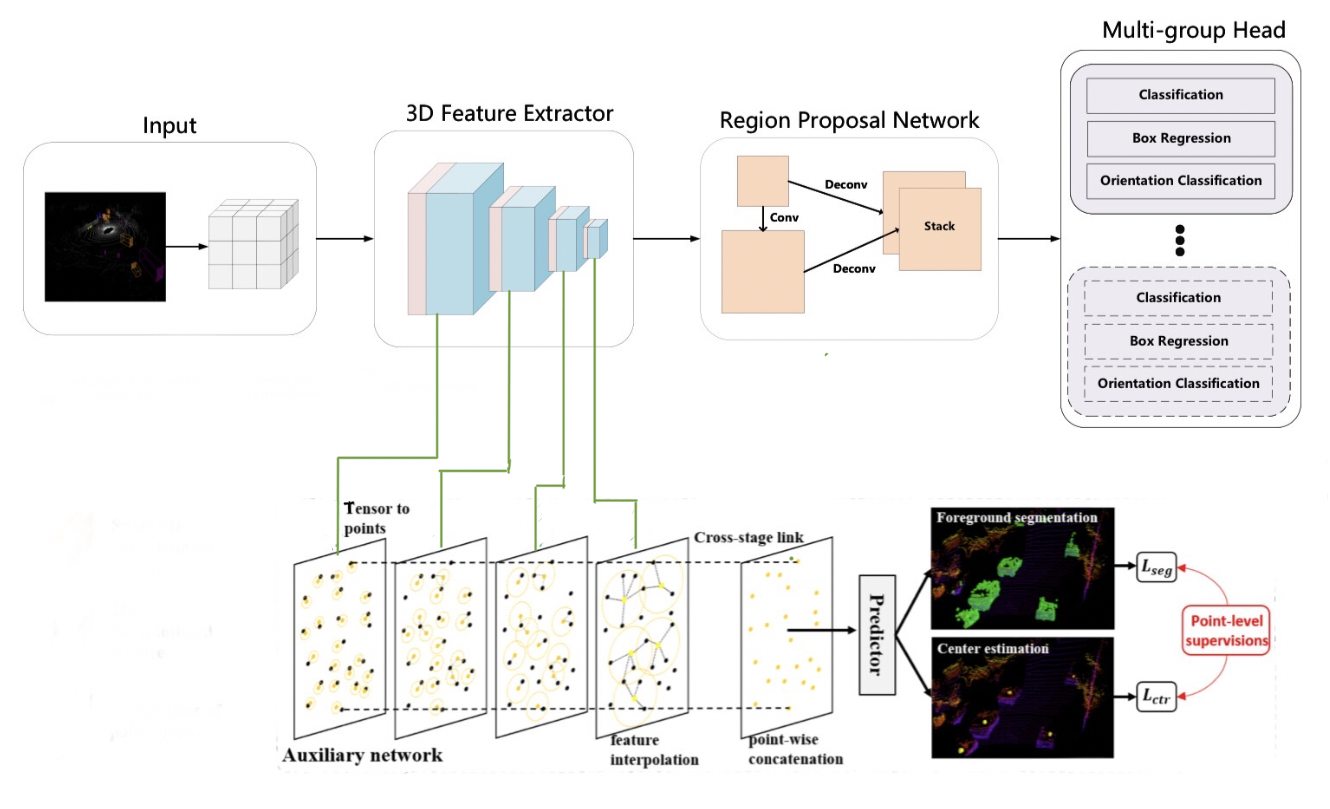}
  \caption{Model architecture}
\end{figure*}

\section{Methodology}
The whole 3D object detection architecture flow diagram is presented in figure 1. The main modules of the architecture include Voxelization Unit, 3D Feature Extractor backbone, Auxiliary Network, Region Proposal Network  and Multi head network. Each of the modules are explained in detail in the following sub sections. In addition, input data sampling and augmentation strategies are explained in the following section.
\subsection{Input and Augmentation}
Each Point cloud frame in NuScenes 3D object dataset has the format $(x,y,z, intensity, ringindex)$. Also each frame is accompanied by a timestamp. As followed in \cite{nuscenes}, 10 lidar sweeps are aggregated together to form a dense point cloud. An example of one such aggregation of lidar sweeps with ground truth boxes is shown in figure [2].The keyframes (samples) are always considered and in between keyframes, 9 other non keyframes (sweeps) are used for aggregation. The input point cloud format used is of the form $(x,y,z, intensity, \Delta t)$ \cite{det3d}. $\Delta t$ is the time lag of non-keyframe with respect to keyframe. 

NuScenes dataset has severe class imbalance problem. To attenuate this problem DS sampling strategy used in CBGS \cite{det3d} is adopted. If examples of a class is less, then more number of duplicates are created to sample from. DS sampling smoothens the class distribution curve. Another common augmentation strategy GT-AUG proposed in SECOND \cite{Second} is used to maintain a database of point clouds of various objects and augment the training samples with it.

VoxelNet\cite{voxelnet} based voxelization strategy is followed. Voxel size of $[0.1m,0.1m,0.2m]$ is used. The point cloud is divided into voxels and each voxel is represented byb the average of all the points present in it.

\subsection{Backbone and detection networks}
The backbone network down samples the voxelized representation of the point cloud into a smaller but semantically strong and dense feature representation. This is achieved using a deep network of sparse 3D convolutions with skip connections. Though the final output feature representation of the backbone is semantically accurate, it lacks spatial accuracy. This is addressed through an auxiliary network proposed in the following section.

An end to end trainable Region Proposal Network used in VoxelNet \cite{voxelnet} is adopted. The RPN consists of series of down sampling layers and concatenation of each down sampled output to get an high resolution feature map.

\subsection{Auxiliary network}
As we progressively down sample in the backbone network, the spatial resolution of the feature maps decrease. So, the features at the boundaries may get merged with background. Also, there is ambiguity in determining the scale and shape of bounding boxes since the feature maps are very sparse.

Auxiliary network solves this by making the backbone network structurally aware. The auxiliary network proposed in SA-SSD \cite{sassd} is adopted. This auxiliary network with point wise supervision is used to guide the intermediate layers of the backbone to extract fine grained structure of the point cloud.Figure [2]

The auxiliary network maps each non zero index of backbone feature to real world coordinates so that each backbone feature can be represented in a point-wise form. This representation is denoted by ${(f_{j},p_{j}) : j = 1,2, ...M }$ where $f$ is the feature vector and $p$ is the point coordinate. Now interpolation is used to generate full resolution feature vector $\tilde{f}$ at the original point cloud coordinates $(p_{i}: i = 1,2,,...N)$. The final representation is of the form ${(\tilde{f_{j}},p_{i}) : i = 1,2, ...N }$. 

The feature vector at each point is calculated by:
\begin{equation}
    \tilde{f_{i}} = \dfrac{\sum_{j=1}^{M}w_{j}(p_{i})f_{j} }{\sum_{j=1}^{M}w_{j}(p_{i})}
\end{equation}
where,
\begin{equation}
w_{j}(p_{i}) = \begin{cases}
\dfrac{1}{||p_{i}-p_{j}||}, & \text{if $ p_j \epsilon N(p_{i})$}\\
0, & \text{otherwise}
\end{cases}
\end{equation}

Here $N(p_{i})$ denotes a ball of region which has a radius of 0.2m,0.4m and 0.8m at different steps. We concatenate the point wise features generated across different down sampling layers and use a shallow fully connected predictor to generate task specific outputs. The two task specific heads are segmentation and center point estimation heads.

There are two auxiliary tasks defined. The first task applies sigmoid function to the output of segmentation head. This helps in calculating foreground background segmentation loss. This loss helps in making the network more boundary aware. The next task estimates the center of each object in the point cloud sample based on the regression head output ($R^{Nx3}$). A center estimation smooth L1 loss is defined in the following sections. This loss helps the network learn the inter object relationships.

Finally, the auxiliary loss is used only for training. It is disabled while testing. So there is no inference time overhead.

\subsection{Class balanced grouping}
We adopt the class balanced grouping strategy \cite{det3d} to address the class imbalance in the nuScenes dataset. The idea is that classes of similar shape or size are easier to learn from the
same task. Classes with similar shapes or sizes share some common features that can compensate for each other during training. The model first recognizes the superclass or group to which an object belongs, and within each group there are object classification, regression and orientation classification heads. Further, in order to balance the instance numbers in each group, major classes are not included as a part of groups. For example, since cars account for 43.7\% of the dataset, adding this to a group will dominate the learning process. 

The 10 classes are thus split into 6 groups: (Car), (Truck, Construction
Vehicle), (Bus, Trailer), (Barrier), (Motorcycle, Bicycle),
(Pedestrian, Traffic Cone).

\subsection{Multi-Group Non-Maximum Suppression}
During inference, in addition to performing Non-Maximum Suppression (NMS) within each group, we also apply NMS across all groups to further reduce the overlapping predictions.  A score threshold of 0.1 and an IoU threshold of 0.3 is used. The NMS technique used also takes into account the predicted yaw angle to compute the IoU.

\subsection{Loss functions}
The three branches in each group are object classification, bounding box regression and orientation classification. Regression without normalization is performed for velocity estimation. Different classes have different anchor dimensions depending on the class mean values. This is adopted from VoxelNet \cite{voxelnet}. The losses used are weighted focal loss, smooth l1 loss and softmax cross-entropy loss for classification, regression and orientation classification respectively. The regression terms are: $(x,y,z,l,w,h,v_x,v_y)$. 

In the auxiliary network, we use focal loss for the foreground segmentation task, and smooth-l1 loss for center estimation. 

\begin{equation}
    \small{L_{seg} = \dfrac{1}{N_{pos}}\sum_{1}^{N}-\alpha(1-\hat{s_i})^{\gamma}\log(\hat{s_i})}
\end{equation}
\begin{equation}
    \hat{s_i}=\begin{cases}
    \tilde{s_i}, & \text{if $s_i=1$}.\\
    1-\tilde{s_i}, & \text{otherwise}.
  \end{cases}
\end{equation}
where $\tilde{s_i}$ denotes the predicted foreground/background probability of each point and $s_i$ is a binary label to indicate
whether a point falls into a ground-truth bounding box. $\alpha$ and $\gamma$ are hyper parameters, and values of 0.25 and 2 are used for the same. 
\begin{equation} 
    L_{ctr} = \dfrac{1}{N_{pos}}\sum_{1}^{N}Smooth-l1(\Delta\tilde{p} - \Delta p).1[s_i=1]
\end{equation}
where $N_{pos}$ is the number of foreground points and $1[.]$ is an indicator function.

The detection and auxiliary tasks are jointly optimized using the following objective: 
\begin{equation}
    L = L_{cls} + \omega L_{box} + \beta L_{orient} + \mu L_{seg} + \lambda L_{ctr}
\end{equation}

\section{Implementation details}
\subsection{Network}
\textit{Voxelization:} The voxel size is $s_x = 0.1, s_y = 0.1, s_z = 0.2$ meters. Point cloud within the range of [-50.4, 50.4], [-51.2, 51.2], [-5.0, 3.0] meters in X, Y, Z axis respectively is considered, and the number of voxels is 1008 x 1024 x 40. 

For the 3D feature extractor, 16, 32, 64, 128 channels of sparse 3D convolution respectively for each block is used. We take outputs from layers 8,12,16 and 18 for the auxiliary network. Each nonzero index of the backbone feature is converted to real-world coordinates based on
the quantization step of the current stage, so that it can be represented in a point-wise form. To generate full-resolution point-wise features, a employ a feature propagation layer \cite{pointnet++} is employed at each stage to interpolate backbone features at the coordinates of original point cloud. For interpolation, the inverse distance weighted average among all the points in a neighboring region is used.

Data augmentation is applied during training by using a random flip in the x-axis. Scaling is performed with a scale factor sampled from [0.95, 1.05]. Rotation is applied around Z axis between [-0.3925, 0.3925] radians and translation is performed in the range [0.2, 0.2, 0.2] m in all the axes.

\subsection{Training}
The model was trained on Nvidia Gtx 1080 GPU. We use adamW \cite{adamw} optimizer together with one-cycle policy with a maximum learning rate of 0.04, with division factor 10. The momentum ranges from 0.95 to 0.85, and a fixed weight decay 0.01 is used to achieved faster convergence. The model is trained for 5 epochs with batch size 4. During inference, top 1000 proposals are retained in each group, then NMS with score threshold of 0.1 is applied, with an IoU threshold of 0.2. A threshold of 80 is set for the number of boxes in each group after NMS.

\section{Experimental Evaluation}

\subsection{Dataset}
The nuScenes dataset \cite{nuscenes} consists of 1000 driving scenes collected in Boston and Singapore. 

For object detection and tracking problems, 23 objects classes are annotated with accurate 3D bounding boxes at 2Hz. Additionally, object-level attributes such as visibility, activity and pose are also annotated. The entire dataset contains 1.4M camera images, 390K LIDAR sweeps, 1.4M RADAR sweeps, and 1.4M object bounding boxes in 40k keyframes. 

\begin{table*}[h!]

\begin{center}
    
\begin{tabular}{||c c c c c c c c||} 
\hline
Method & mAP & mATE & mASE & mAOE & mAVE & mAAE & NDS \\ [0.5ex] 
\hline\hline
CBGS & 18.96 & 0.57 & 0.36 &  1.22 & 1.40 & 0.32 & 26.8 \\ 
\hline
BS-CBGS ($\mu = 1, \lambda = 0$) & 16.54 & 0.62 & 0.30 &  \textbf{1.087} & 2.79 & 0.37 & 25.1\\ [1ex] 
\hline
SA-CBGS ($\mu = 1, \lambda = 2$) & 15.56 & 0.588 & 0.314 &  1.23 & 1.73 & \textbf{0.223} & 26.5 \\ [1ex] 
\hline
SA-CBGS ($\mu = 0.5, \lambda = 1$) & 20.10 & 0.56 & 0.37 &  1.26 & 1.48 & 0.32 & 27.37\\ [1ex] 
\hline
SA-CBGS ($\mu = 2, \lambda = 4$) & \textbf{20.67} & \textbf{0.527} & \textbf{0.286} &  1.10 & 2.23 & 0.284 & \textbf{29.36}\\ [1ex] 
\hline
SA-CBGS+NMS ($\mu = 2, \lambda = 4, $ t=0.1) & 19.56 & 0.506 & 0.283 &  1.08 & \textbf{2.15} & 0.27 & 29.08\\ [1ex] 
\hline

\end{tabular}
\caption{\label{tab:table-name} Evaluation Metrics-10\% data}
\end{center}
\end{table*}

\begin{table*}[h!]

\begin{center}
    
\begin{tabular}{||c c c c c c c c||} 
\hline
Method & mAP & mATE & mASE & mAOE & mAVE & mAAE & NDS \\ [0.5ex] 
\hline\hline
CBGS-40GB & 26.05 & \textbf{0.48} & 0.29 &  1.24 & \textbf{0.59} & \textbf{0.23} & \textbf{36.95} \\ [1ex] 
\hline
SA-CBGS-40GB & \textbf{26.37} & 0.49 & 0.29 &  1.24 & 0.63 & 0.25 & 36.41 \\ [1ex] 
\hline

\end{tabular}
\caption{\label{tab:table-name} Evaluation Metrics-33\% data}
\end{center}

\end{table*}

\begin{table*}[h!]
\begin{center}
 \begin{tabular}{||c c c c c c c c||}
 \hline
 Method & Car & Truck & Construction Vehicle & Bus & Barrier & Pedestrian & Traffic Cone \\ [0.5ex] 
 \hline\hline
 CBGS & 55.79 & 8.80 & 4.10 & 10.76 & 16.17 & 53.90 & 31.21 \\ 
 \hline
BS-CBGS ($\mu = 1, \lambda = 0$) & 43.75 & 5.13 & 3.35 & 5.87 & \textbf{19.94} & 49.89 & \textbf{34.69}\\ [1ex] 
 \hline
 SA-CBGS($\mu = 1, \lambda = 2$) & 46.7 & 5.79 & 2.60 &  10.74 & 13.98 & 45.03 & 30.01\\ [1ex] 
 \hline
 SA-CBGS ($\mu = 0.5, \lambda = 1$) & 52.85 & \textbf{14.54} & 7.37 &  \textbf{20.06} & 17.46 & 52.54 & \textbf{34.69}\\ [1ex] 
 \hline
 SA-CBGS ($\mu = 2, \lambda = 4$) & \textbf{56.30} & 12.8 & 10.0 & 16.27 & 19.83 & \textbf{56.70} & 29.6\\ [1ex] 
 \hline
 SA-CBGS+NMS ($\mu = 2, \lambda = 4$, t=0.1) & 52 & 11.2 & \textbf{10.12} & 15.4 & 16.57 & 56.29 & 28.89\\ [1ex] 
 \hline
 
\end{tabular}
\caption{\label{tab:table-name} mAP values for different classes}
\end{center}
\end{table*}

\begin{figure*}[h!]
  \centering
  \subfigure[PR curve for car on CBGS]{\includegraphics[width=7cm, height=5cm, scale=0.3]{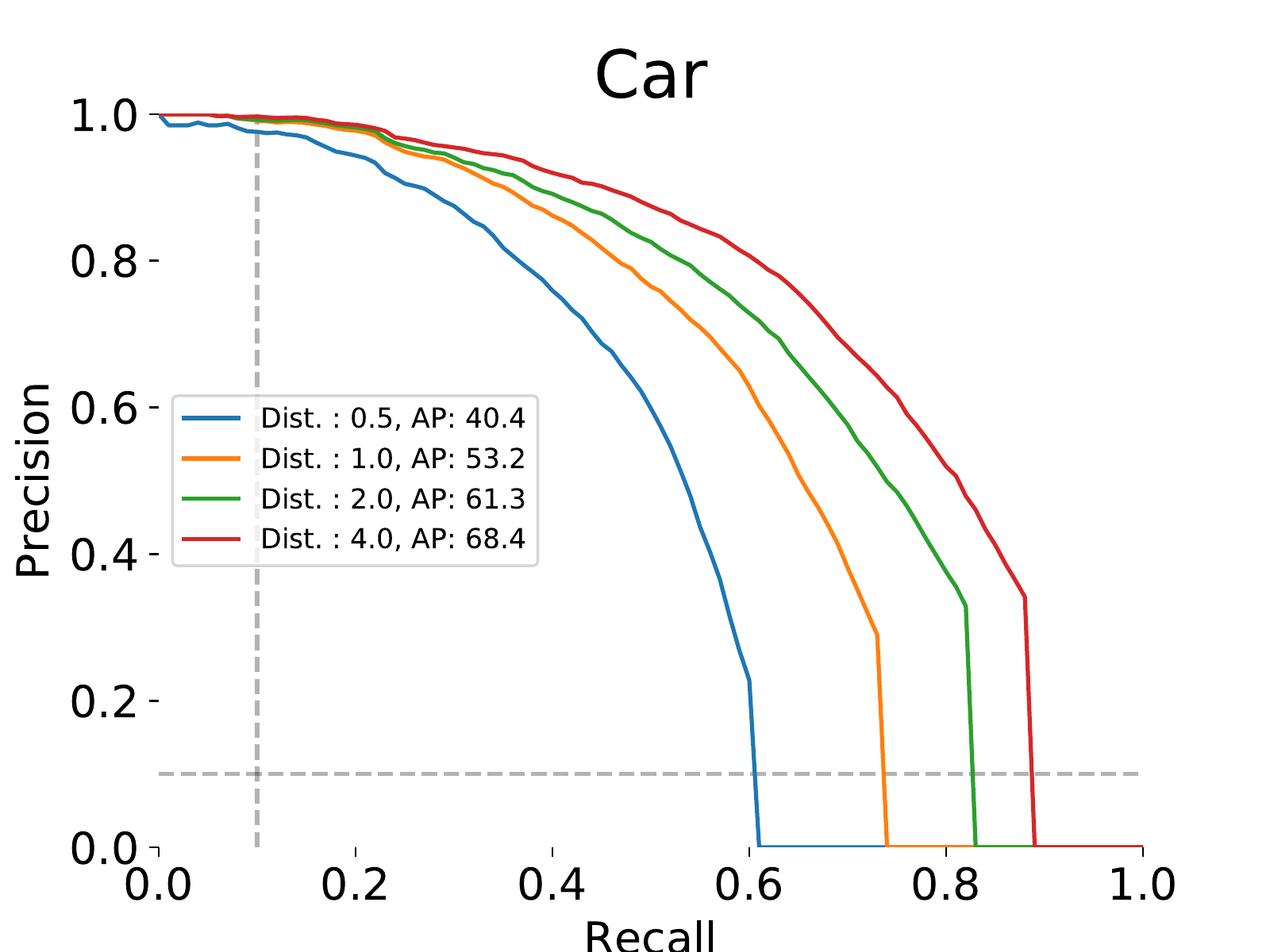}}\quad
  \subfigure[PR curve for car on SA-CBGS]{\includegraphics[width=7cm, height=5cm, scale=0.3]{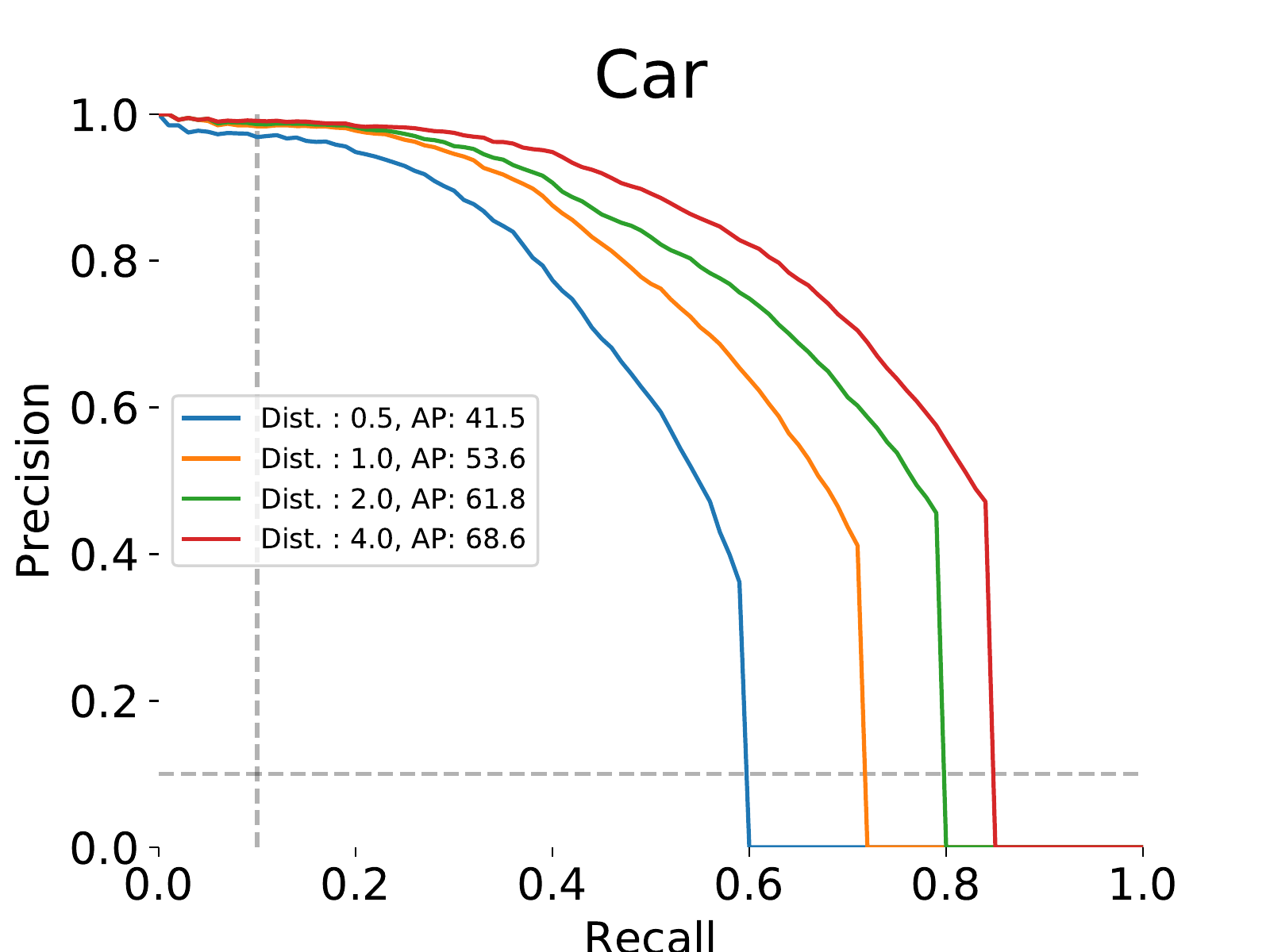}}
  \caption{PR curves for class car for CBGS and SA-CBGS}
\end{figure*}

\subsection{Metrics}

The metrics for the nuScenes detection task will be described below. The final score, also known as the nuScenes Detection Score is a weighted sum of the mean Average Precision (mAP) and the True Positive (TP) metrics. 

\subsubsection{Average Precision Metric}
For the Average Precision metric, a match is defined based on the 2D center distance on the ground plane. This is chosen in place of intersection over union (IoU) in order to account for objects with relatively smaller footprints, like bikes and pedestrian. If such objects are detected, a small error in translation can lead to an IoU of 0, which is undesirable. 

The AP is computed as the normalized area under the precision recall curve for precision and recall over 10\%. The mAP is obtained by averaging over the AP obtained for the matching thresholds of D = {0.5, 1, 2, 3} metres and a set of classes C. \vspace{-.3cm}

$$ mAP = \frac{1}{|C||D|} \Sigma_{c \in C} \Sigma_{d \in D} AP_{c,d}$$

\subsubsection{True Positive Metrics}
For every prediction that matched with a ground truth box, a set of True Positive metrics are measured. The metrics are computed using the matching distance threshold of d = 2 metres. \vspace{-0.2cm}
\begin{itemize}[noitemsep]
  
    \item Average Translation Error (ATE): 2D Euclidean distance between the predicted and ground truth centers (units in meters).
    \item Average Scale Error: Error in 3D IoU after alignment of orientation and translation. Calculated as (1 - IoU).
    \item Average Orientation Error: Yaw angle difference between the prediction and ground truth (units in radians).
    \item Average Velocity Error: L2 norm of the difference in velocities (units in m/s).
    \item Average Attribute Error: 1 - acc, where acc is the attribute classification accuracy for attributes such as sitting or standing.
\end{itemize}

For each TP metric, the mean TP metric (mTP) is computed over all classes,
\vspace{-0.3cm}
$$ mTP = \frac{1}{|C|} \Sigma_{c \in C} TP_c $$

\subsubsection{nuScenes Detection Score}
As the mAP does not capture all aspects of the nuScenes detection tasks, like velocity and attribute estimation, the nuScenes Detection Score is used as the final score. It is computed by consolidating the mAP and TP metrics into a scalar score,
\vspace{-0.2cm}
$$ NDS = \frac{1}{10}[5 mAP + \Sigma_{mTP \in TP} (1 - min(1, mTP))] $$

where the mAP is the mean Average Precision, and TP is the set of the 5 mean True Positive metrics.



    




\subsection{Experiments}
A series of experiments were conducted to analyze the effect of incorporating auxiliary network into CBGS architecture\cite{det3d}. First a model based on CBGS \cite{det3d} was trained to get baseline model to compare with. Then, loss weights for center estimation ($\mu$) and foreground-background estimation ($\lambda$) suggested by SA-SSD model \cite{sassd} was used to train the proposed SA-CBGS network. Then, a range of experiments for various values of $\mu$ and $\lambda$ were tried to tune the loss weights. Also, an experiment using only the foreground-background segmentation loss was tried to evaluate the spatial localization of the trained model. All these experiments were carried out with 10\% of the nuScenes dataset. Another set of experiments evaluated the baseline and SA-CBGS model on 33\% of the nuScenes dataset. Finally, inter class Non Maximal Suppression was applied on a trained SA-CBGS model with weights $\lambda$ = 4 and $\mu$ = 2.

\subsection{Results}
Table 1 contains the the evaluation metrics for the experiments described above. Upon fine-tuning the weights for computing the auxiliary loss, the weights $\mu = 2, \lambda = 4$ outperformed the baseline by 1.71 points. The SA-CBGS networks with any other auxiliary loss weights failed to match the metrics of the baseline. The reasoning for this could be that the contribution of the loss in the weight updates was not high enough to have an impact. 
It can also be seen that the mATE and mASE have significantly reduced as these can be directly correlated to the auxiliary tasks of foreground segmentation and center point estimation. The foreground segmentation task makes the features more aware of the boundary, and therefore the scale of the objects. The center point estimation task influences the feature maps to reduce the translation error in the centers of the predicted bounding boxes.

As the best weight values were found to be $\mu=2, \lambda=4$, the rest of the experiments were conducted with those as the auxiliary loss weight values. Inter-group NMS was applied as a post processing step, and although it seems to have visually improved the results the evaluation metrics have not showed improvement. This is because the threshold on IoU could lead to False Negatives, hence deteriorating performance.

The mAP values for each category are compared in Table 3. Although there is no particular trend among the results of the different experiments, the structurally aware models in general seem to outperform the baseline.  

Another experiment with more amount of data (33\%) was performed with loss weights $\mu=1, \lambda=2$. The results for this are reported in Table 2. Here the mAP values for both the baseline and SA-CBGS models were comparable. This can be attributed to using of different non optimal loss weights which was figured out in the previous experiments. Also the mAP values obtained with 33\% of the data were significantly more than the best performing model on 10\% of the data. This suggests us that the previous model on less amount of data was under fitting and has scope for improvement. 

The precision recall curves for the Car category computed using various distance values can be seen in Figure 4. The PR curve for for SA-CBGS is reported for the best performing model with $\mu = 2, \lambda = 4$. The area under the curve is slightly higher for the SA-CBGS model, which explains the higher mAP compared to the baseline.  Examples of qualitative detection results can be seen in Figures 5, and 6. The edge with a line attached in the bounding box indicates the vehicle’s front. Figure 7 contains the detection results after applying NMS on the predicted boxes. The green colored boxes are the ground truth 3D object boxes and the blue boxes are the predicted 3D object boxes.

\section{Conclusion}
We proposed to use an auxiliary network to guide the features learned by the 3D extractor to be more aware of the structure information of the 3D objects. As the feature maps are more aware of the structure, the object detection performance outperforms the baseline significantly in terms of multiple evaluation metrics including the nuScenes Detection Score. 
\begin{figure*}[!htp]
  \centering
  \subfigure[Qualitative results of CBGS - scene 1]{\includegraphics[scale=0.28]{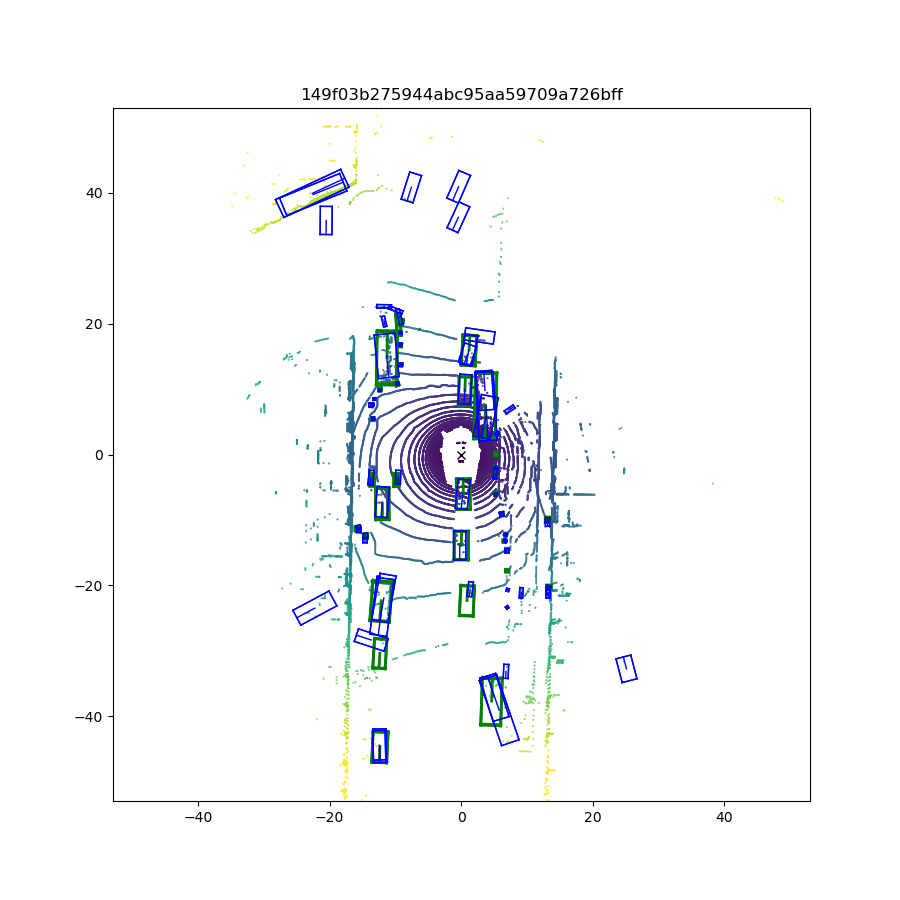}}\hspace{0.1\textwidth}
  \subfigure[Qualitative results of SA-CBGS - scene 1]{\includegraphics[scale=0.28]{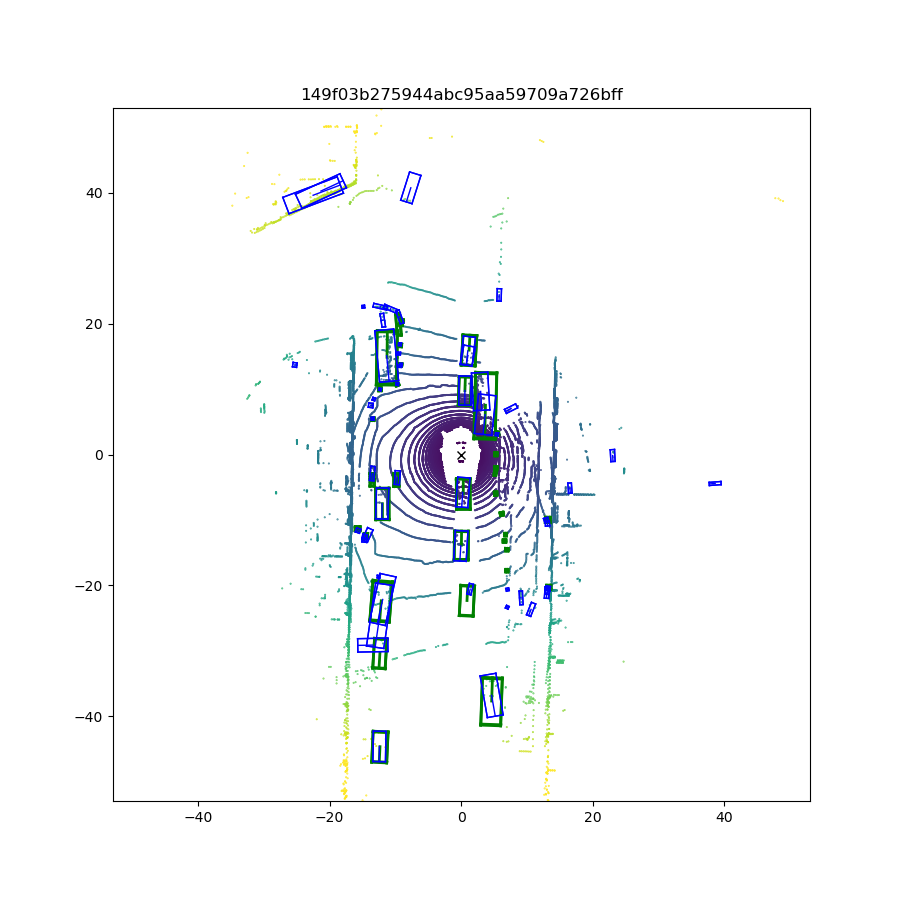}}
  \caption{Qualitative results - scene 1}
\end{figure*}
\begin{figure*}[!htp]
 \centering
  \subfigure[Qualitative results of CBGS - scene 2]{\includegraphics[scale=0.28]{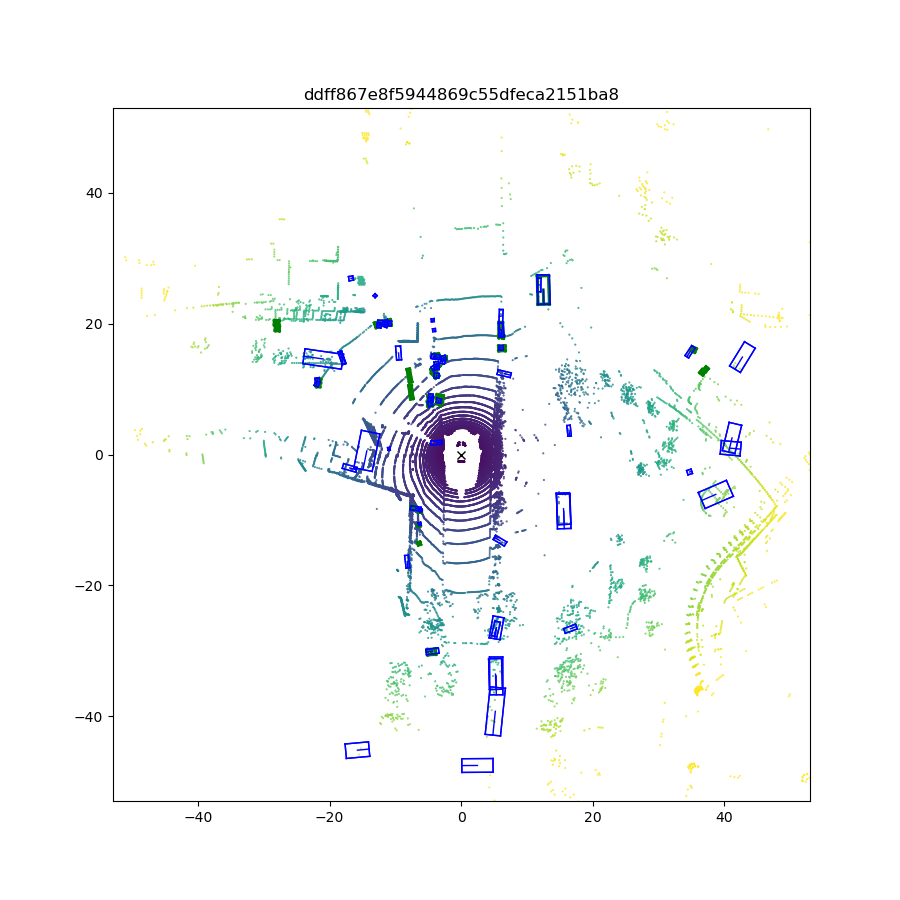}}\hspace{0.1\textwidth}
  \subfigure[Qualitative results of SA-CBGS - scene 2]{\includegraphics[scale=0.28]{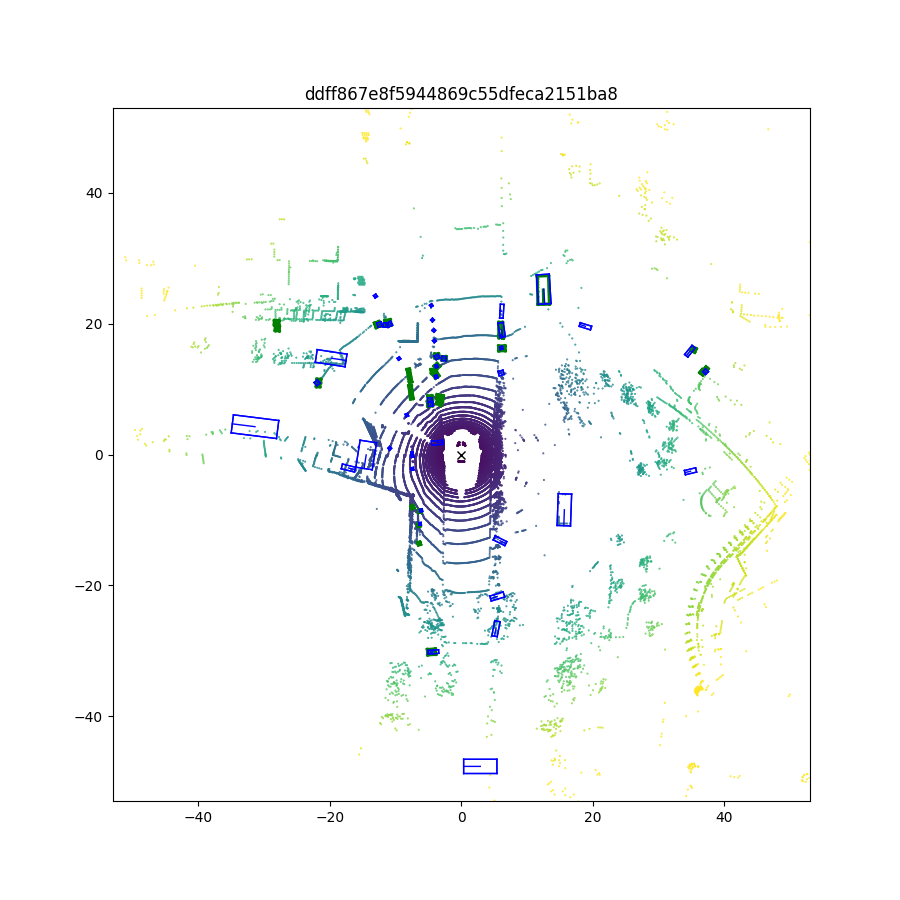}}
  \caption{Qualitative results - scene 2}
\end{figure*}

\begin{figure*}[!htp]
  \centering
  \subfigure[Original detection]{\includegraphics[scale=0.28]{pictures/scene1_best.png}}\hspace{0.1\textwidth}
  \subfigure[Detection with NMS]{\includegraphics[scale=0.28]{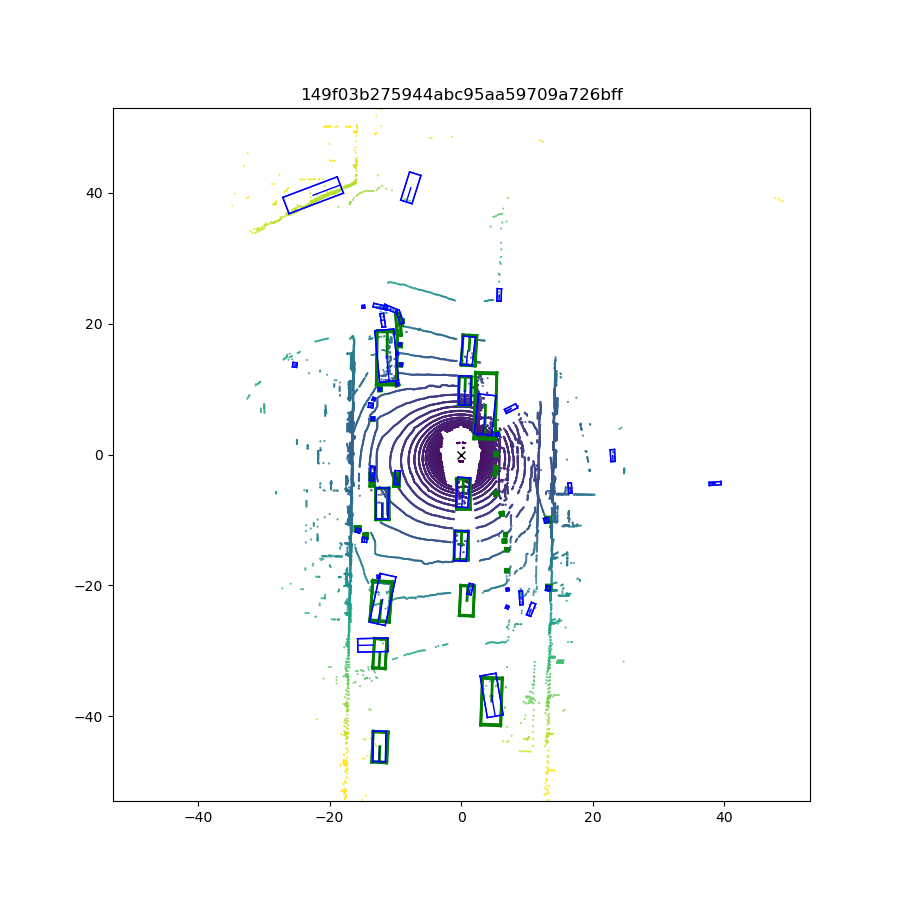}}
\end{figure*}
\begin{figure*}[!htp]
  \centering
  \subfigure[Original detection]{\includegraphics[scale=0.28]{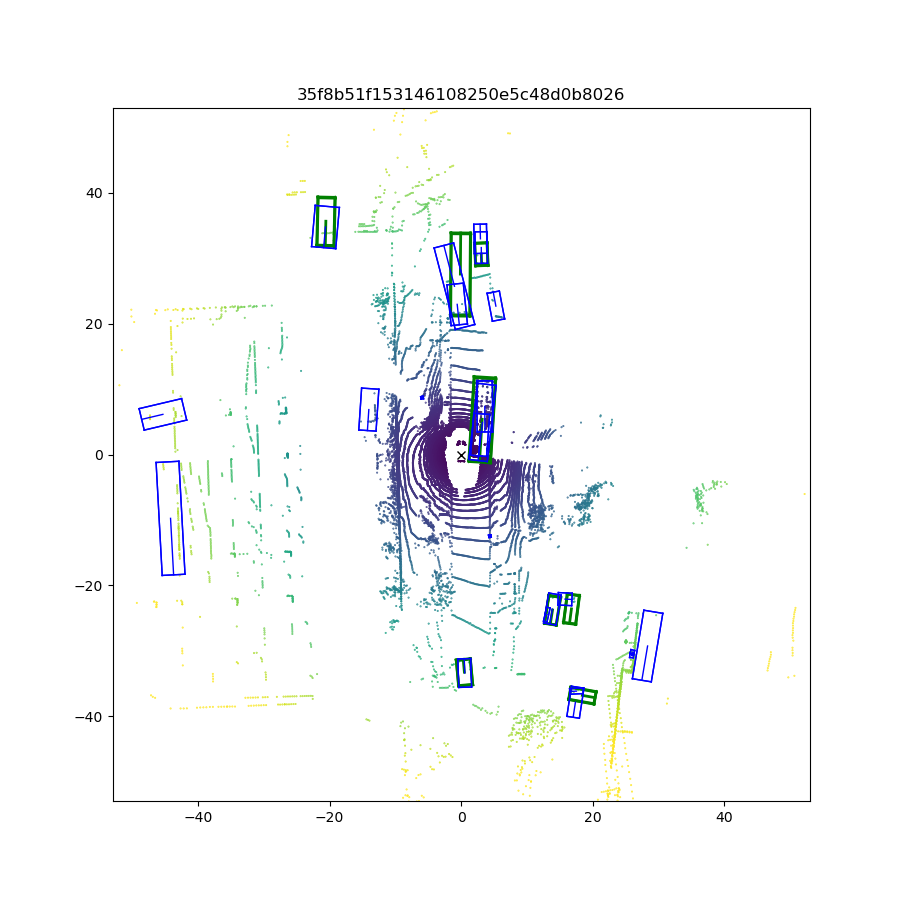}}\hspace{0.1\textwidth}
  \subfigure[Detection with NMS]{\includegraphics[scale=0.28]{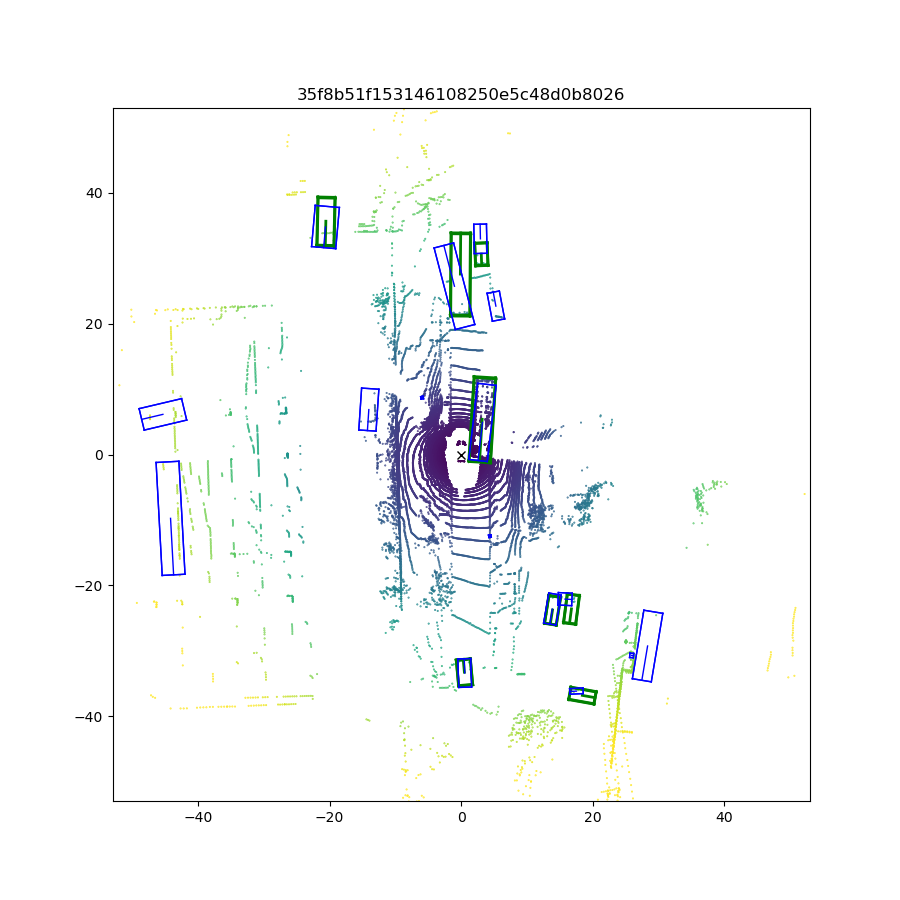}}
  \caption{Effect of multi-group NMS}
\end{figure*}
{\small
\bibliographystyle{ieee_fullname}
\bibliography{egbib}
}
\newpage
\end{document}